\documentclass[sigconf]{acmart}

\AtBeginDocument{%
  \providecommand\BibTeX{{%
    \normalfont B\kern-0.5em{\scshape i\kern-0.25em b}\kern-0.8em\TeX}}}

\setcopyright{acmcopyright}
\copyrightyear{2018}
\acmYear{2018}
\acmDOI{XXXXXXX.XXXXXXX}

\acmConference[Conference acronym 'XX]{Make sure to enter the correct
  conference title from your rights confirmation emai}{June 03--05,
  2018}{Woodstock, NY}
\acmPrice{15.00}
\acmISBN{978-1-4503-XXXX-X/18/06}

\usepackage{multirow}
\usepackage{bm}
\usepackage{subcaption}
\usepackage{graphicx}
\usepackage{color, soul}
\usepackage{enumitem}
\usepackage{amsmath,amsthm,mathtools}
\usepackage{colortbl}
\definecolor{mygray}{gray}{0.9}
\usepackage{wrapfig}
\usepackage{changepage}
\usepackage{balance}

\begin{document}

\title{GiGaMAE: Generalizable Graph Masked Autoencoder via \\ Collaborative Latent Space Reconstruction}

\author{Yucheng Shi}
\orcid{0009-0007-4192-1315}
\affiliation{%
  \institution{University of Georgia}
  \city{Athens}
  \state{Georgia}
  \country{USA}
}
\email{yucheng.shi@uga.edu}

\author{Yushun Dong}
\orcid{0000-0001-7504-6159}
\affiliation{%
  \institution{University of Virginia}
  \city{Charlottesville}
  \state{Virginia}
  \country{USA}
}
\email{yd6eb@virginia.edu}

\author{Qiaoyu Tan}
\orcid{0000-0001-8999-968X}
\affiliation{%
  \institution{New York University Shanghai}
  \city{Shanghai}
  \country{China}
}
\email{qiaoyu.tan@nyu.edu}


\author{Jundong Li}
\orcid{0000-0002-1878-817X}
\affiliation{%
  \institution{University of Virginia}
  \city{Charlottesville}
  \state{Virginia}
  \country{USA}
}
\email{jundong@virginia.edu}

\author{Ninghao Liu}
\orcid{0000-0002-9170-2424}
\affiliation{%
  \institution{University of Georgia}
  \city{Athens}
  \state{Georgia}
  \country{USA}
  }
\email{ninghao.liu@uga.edu}


\begin{abstract}
Self-supervised learning with masked autoencoders has recently gained popularity for its ability to produce effective image or textual representations, which can be applied to various downstream tasks without retraining. However, we observe that the current masked autoencoder models lack good generalization ability on graph data. To tackle this issue, we propose a novel graph masked autoencoder framework called GiGaMAE.
Different from existing masked autoencoders that learn node presentations by explicitly reconstructing the original graph components (e.g., features or edges), in this paper, we propose to collaboratively reconstruct informative and integrated latent embeddings. By considering embeddings encompassing graph topology and attribute information as reconstruction targets, our model could capture more generalized and comprehensive knowledge.
Furthermore, we introduce a mutual information based reconstruction loss that enables the effective reconstruction of multiple targets. This learning objective allows us to differentiate between the exclusive knowledge learned from a single target and common knowledge shared by multiple targets. We evaluate our method on three downstream tasks with seven datasets as benchmarks. Extensive experiments demonstrate the superiority of GiGaMAE against state-of-the-art baselines.
We hope our results will shed light on the design of foundation models on graph-structured data. Our code is available at: \url{https://github.com/sycny/GiGaMAE}.
\end{abstract}

\ccsdesc[500]{Computing methodologies~Machine learning}
\ccsdesc[500]{Computing methodologies~Artificial intelligence}

\keywords{Self-supervised Learning, Graph Mining, Masked Autoencoder.}

\maketitle

\section{Introduction}
\label{introduction}
 
\begin{figure}[t]
\centering
\includegraphics[width=0.47\textwidth]{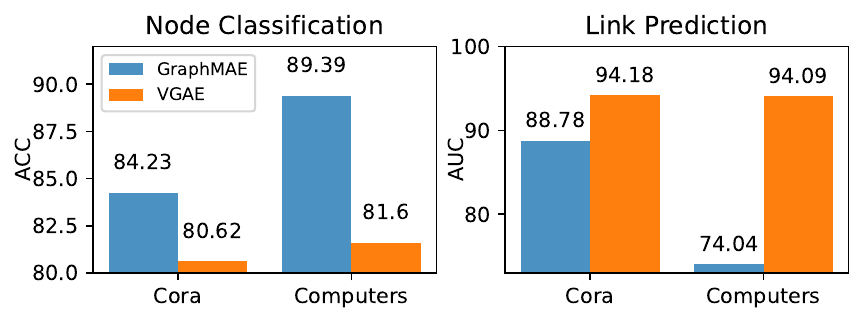}
\vspace{-12pt}
\caption{Preliminary experiments on two downstream tasks (node classification and link prediction) with GraphMAE~\cite{hou2022graphmae} and VGAE~\cite{kipf2016variational} on Cora and Computers datasets.} \label{pre_exper}
\vspace{-10pt}
\end{figure}

Self-supervised generative models, exemplified by MAE~\cite{he2022masked} and BERT~\cite{devlin2018bert}, have demonstrated remarkable performance in acquiring generalizable representations in various domains such as computer vision~\cite{kirillov2023segment, wei2022masked} and natural language processing~\cite{zheng2021rethinking}. Such representations offer the advantage of being easily adaptable to diverse downstream tasks. In graph domains, generalizable representations also hold significant value in real applications, such as social network platforms where we may want to conduct recommendation~\cite{fan2019graph, tan2021sparse, tan2021dynamic, zhou2021temporal} (link prediction), community detection~\cite{hu2020going, tan2019deep} (node clustering) and malicious account detection~\cite{cai2021structural, guan2023xgbd, zhou2022unseen} (node classification) simultaneously. Thus, generalizable node representations are desirable.

However, we have observed that it is challenging for existing self-supervised generative models on graphs to meet the above expectation.
To assess the generalization capacity, we conduct a pilot study on two representative models (VGAE~\cite{kipf2016variational} and GraphMAE~\cite{hou2022graphmae}) in Figure~\ref{pre_exper}. We observe that the recently proposed GraphMAE shows good performance in node classification, but it lags behind traditional VGAE in link prediction. 
Moreover, neither model consistently achieves satisfactory results on both tasks.
We have similar observations on other graph generative models as well (e.g., GAE~\cite{kipf2016variational}, Marginalized GAE~\cite{wang2017mgae}, GATE~\cite{salehi2019graph}, and S2GAE~\cite{tan2023s2gae}), and in other tasks (e.g., node clustering).
Based on these observations, we argue that the well-established graph generative methods usually fail to exhibit desirable generalization capabilities across tasks.
The limited generalization ability of the model necessitates additional efforts for training on different downstream tasks, which can be time-consuming in practice~\cite{thakoor2021large}.

To understand the generalization issue above, we analyze the state-of-the-art graph generative models and identify the key obstacles. These models typically follow the design of auto-encoding frameworks~\cite{he2022masked} consisting of \textit{an encoder} that learns to map the input graph~\cite{hou2022graphmae, kipf2016variational} into latent representations, and \textit{a decoder} that reconstructs either the observed edges~\cite{kipf2016variational} or raw node features~\cite{hou2022graphmae,wang2017mgae} from the latent space. Despite their simplicity and popularity, we argue that such a naive reconstruction task is sub-optimal for graph representation learning. This is mainly because graphs are heterogeneous data, containing multiple modalities of information (node features and graph topology). Simply reconstructing a single modality only captures limited aspects of information in the learned representations~\cite{huang2021makes}. 
Therefore, a model achieves good performance in a specific downstream task if the corresponding graph information is encoded in embeddings.
For example, preserving local information such as node features (e.g., GraphMAE) could benefit node classification, while learning graph structural information results in embeddings that are effective for link prediction or clustering.
This motivates us to design a more \textbf{comprehensive self-supervised reconstruction objective} to enhance the generalization ability of representations.

However, this is nontrivial due to the following challenges.
\textbf{(i) Graph reconstruction incompatibility.} 
  Graph topology and attributes possess distinct characteristics. Graph edges are discrete and sparse, while node features are continuous and often high-dimensional. This fundamental incompatibility makes it difficult to reconstruct both edges and features simultaneously. Furthermore, previous research has shown that directly reconstructing both edges and features can adversely impact model performance~\cite{hou2022graphmae}. 
\textbf{(ii) Limitations of existing learning objectives.}
  Current learning objectives are predominantly designed for reconstructing a single modality and do not account for the joint reconstruction of both topology and attribute information. For example, Mean Square Error (MSE) is employed for feature reconstruction tasks, while Binary Cross Entropy (BCE) is used for link prediction tasks. A straightforward approach would be to combine these objectives. However, such optimization goals impose conflicting requirements on the reconstruction model, leading to gradient conflicts~\cite{yu2020gradient,lin2021closer} and resulting in sub-optimal performance.

In this paper, we propose a novel self-supervised generative model on graphs, called GiGaMAE (\underline{G}eneral\underline{i}zable \underline{G}r\underline{a}ph \underline{M}asked \underline{A}uto\underline{E}ncoder), to tackle the above challenges. For challenge (i), instead of directly recovering edges and node features, we map information into a homogeneous latent space for reconstruction. To be more concrete, we use embeddings from multiple external models (e.g., node2vec~\cite{grover2016node2vec} and PCA~\cite{abdi2010principal}) as the reconstruction targets. 
These targets encompass diverse information and can be reconstructed in a unified way, which facilitates the learning of generalizable representation.
For challenge (ii), we leverage the Infomax principle~\cite{tschannen2019mutual} and design a mutual information based reconstruction loss. This loss explicitly captures the shared information between graph topology and node attributes, and distinguishes the distinctive information from different sources. In our framework, we prioritize learning from the shared information since it contains more underlying knowledge, thus enhancing generalization capabilities. We evaluate our framework on various tasks, including node classification, node clustering, and link prediction. 
The main contributions of this paper are summarized below:
\begin{itemize}[topsep=3mm]
\item We investigate how to enhance the generalization capability of self-supervised graph generative models, and propose a novel approach called GiGaMAE by reconstructing graph information in the latent space. 
\item We propose a novel self-supervised reconstruction loss. This loss effectively characterizes, balances, and integrates both shared and distinct information across multiple collaborative reconstruction targets.
\item We conduct extensive experiments on seven benchmark datasets to evaluate the generalization ability of GiGaMAE. Empirical results demonstrate the superiority of our proposed method across three critical graph analysis tasks compared to state-of-the-art baselines. 
\end{itemize}

\section{Preliminaries}

\subsection{Mutual Information (MI)}
\label{Mutual_Information}
We resort to mutual information (MI)~\cite{shannon1948mathematical, shi2023engage} for the learning objective design in our paper since it can measure the dependency relationship between variables~\cite{tishby2000information,gabrie2018entropy}. 

\paragraph{Mutual Information}
The mutual information between two variables $X_1,X_2$ is defined as:
\begin{equation}
 \begin{split}
    &I(X_1;X_2) =KL(P_{(X_1,X_2)}||P_{X_1}P_{X_2}) \\
    = &\sum_{x_1 \in X_1} \sum_{x_2 \in X_2} P_{(X_1, X_2)}(x_1, x_2) \log \left(\frac{P_{(X_1, X_2)}(x_1, x_2)}{P_{X_1}(x_1) P_{X_2}(x_2)}\right),
 \end{split}
\end{equation}
where $P_{X_1}$ and $P_{X_2}$ denote the marginal distributions of $X_1$ and $X_2$, $P_{(X_1,X_2)}$ means the joint distribution. $KL(\cdot||\cdot)$ is the Kullback–Leibler divergence~\cite{csiszar1975divergence}.
Multivariate mutual information is a more general definition of dependency measurement if there are more than two variables. The multivariate mutual information between $n\geq3$ variables $X_1, X_2,...,X_n$ is defined as below:
\begin{equation}
\label{interaction}
    I(X_1;...;X_n) =I(X_1;...;X_{n-1})- I(X_1;...;X_{n-1}|X_{n}),
\end{equation}
where $I(X_1;...;X_{n-1}|X_{n})$ is the conditional mutual information given the variable $X_{n}$.

\paragraph{Chain Rules for Mutual Information}
Specifically, given three random variables $X_1$, $X_2$, and $X_3$, two chain rules for mutual information are listed below:
\begin{equation}
\label{chain1}
    I(X_1; X_2, X_3) = I(X_1; X_3) + I(X_1 ; X_2|X_3),
\end{equation}
\begin{equation}
\label{chain2}
    I(X_1; X_2; X_3) = I(X_1; X_2) + I(X_1; X_3) - I(X_1 ; X_2, X_3).
\end{equation}
The proof for Equation~(\ref{chain1})$\sim$(\ref{chain2}) is provided in the Appendix.

\begin{figure*}[h]
\centering
\includegraphics[width=0.85\textwidth]{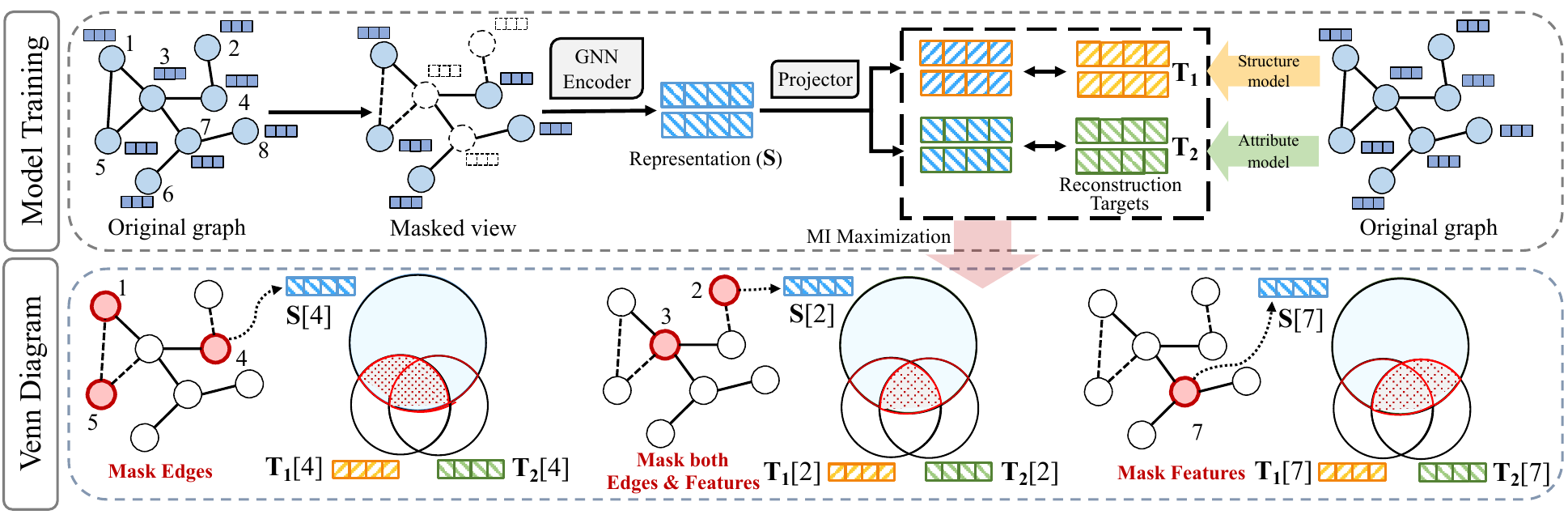}
\vspace{-10pt}
\caption{The framework of GiGaMAE. Top: The training pipeline. Bottom: We learn different information for nodes with different types of masking. The red shadow region in the Venn Diagram highlights the information to be learned by $\mathbf{S}[i]$.} \label{design}
\vspace{-5pt}
\end{figure*}

\subsection{Mutual Information Estimation}
It is difficult to directly compute MI, because the input distribution is usually unknown. Various methods have been proposed to estimate MI~\cite{poole2019variational}. In this paper, we select InfoNCE~\cite{oord2018representation} as the estimator, which has been empirically proven to be effective in various scenarios~\cite{tschannen2019mutual,chen2020simple}. Given a pair of node representations $(p_{i}$, $q_{i})$ as input, the InfoNCE loss $ \ell^{\mathcal{D}}{(p_i, q_i)}$ can be defined as:
\begin{equation}
\label{infornce1}
    \ell^{\mathcal{D}}{(p_i, q_i)}=\log \frac{\mathcal{D}\left(p_{i}, q_{i}\right)}{\sum_{i'=1}^{N} \mathcal{D}(p_{i}, q_{i'}) + \sum_{i'=1}^{N} \mathcal{D}(p_{i}, p_{i'}) - \mathcal{D}(p_{i}, p_{i})},
\end{equation}
where $N$ denotes the number of negative examples, and $\mathcal{D}(\cdot, \cdot)$ denotes the discriminator function. Intuitively, $\mathcal{D}(\cdot, \cdot)$ assigns high values to positive pairs and low values to all other pairs~\cite{velivckovic2018deep, tschannen2019mutual}. In this paper, we define positive pairs as distinct latent representations of the same node in a graph, whereas negative examples are all the representations of other nodes within the same graph. 
Then, the overall objective for set $\mathcal{P}$ and $\mathcal{Q}$ (they both contain $N'$ node representations) is defined as:
\begin{equation}
\label{infornce2}
    \mathcal{L}_{\mathcal{D}}(\mathcal{P},\mathcal{Q}) = \frac{1}{2N'} \sum_{i=1}^{N'}[\ell^{\mathcal{D}}{(p_{i}, q_{i})}+\ell^{\mathcal{D}}{(q_{i}, p_{i})}],
\end{equation}
where $p_i \in \mathcal{P}, q_i \in \mathcal{Q}$. The loss $\mathcal{L}_{\mathcal{D}}(\mathcal{P},\mathcal{Q})$ is a lower bound of mutual information $I(\mathcal{P},\mathcal{Q})$~\cite{poole2019variational}. Specifically, we can maximize the mutual information by maximizing its corresponding InfoNCE loss.

\subsection{Problem Definition}
We define a graph as $\mathcal{G} = \left \{ \mathcal{V}, \textbf{A}, \textbf{X} \right \}$, where $\mathcal{V}$ denotes the node set, $\textbf{A} \in \left\{0,1\right\}^{|\mathcal{V}|\times |\mathcal{V}|}$ is the adjacency matrix, and $\textbf{X}\in \mathbb{R}^{|{\mathcal{V}}|\times D}$ is the feature matrix with $D$-dimensional features for each node. Given a graph $\mathcal{G}$ as input, our framework aims to obtain generalizable node representations $\mathbf{Z} \in \mathbb{R}^{|\mathcal{V}|\times d}$, where $d$ is the latent dimension. $\mathbf{z}_i = \mathbf{Z}[i,:] $ denotes the representation of node $v_i \in \mathcal{V}$.

\newtheorem{problem}{Problem}
\begin{problem}
\textbf{Learning Generalizable Graph Representations.}
Given an input graph $\mathcal{G}$, our goal is to pre-train a generative model to learn node representations $\mathbf{Z}$ that demonstrate consistently competitive performance across commonly studied downstream tasks, including node classification, node clustering, and link prediction.
\end{problem}

\section{Methodology}

We now present GiGaMAE framework. In Section~\ref{GiGaMAEcomponents}, we provide an overview of our model as a graph masked autoencoder. In Section~\ref{re_loss}, we introduce the details of the loss function design for training the model. Finally, in Section~\ref{teacher}, we discuss different candidates as reconstruction targets in the masked autoencoder.

\subsection{Framework Design}
\label{GiGaMAEcomponents}

Our GiGaMAE framework is based on a graph masked autoencoder, as depicted in Figure~\ref{design}. Unlike previous works that directly reconstruct the original graph information (e.g., node features and graph topology), we employ the output of other graph representation models as the reconstruction targets. The details are given below.

\subsubsection{Graph Masked Autoencoder.}
\label{graph_generative_model}
A typical graph masked autoencoder consists of three components: a graph augmenter $f^{aug}(\cdot)$, an encoder $f^{enc}(\cdot)$, and a decoder $f^{dec}(\cdot)$. The original graph $\mathcal{G}$ serves as the input and undergoes masking to become $\mathcal{G}' = f^{aug}(\mathcal{G})$ through edge masking~\cite{tan2023s2gae} or feature masking~\cite{wang2017mgae}.  
The encoder $f^{enc}(\cdot)$ takes $\mathcal{G}'$ as input and encodes the nodes into representations $\mathbf{Z}$. Commonly used GNN architectures, such as GAT~\cite{velivckovic2017graph} and GCN~\cite{kipf2016semi, shi2023chatgraph}, can be applied for the encoder. Finally, the decoder $f^{dec}(\cdot)$ reconstructs graph components, such as edges or features, from the latent representations $\mathbf{Z}$. The learning objective is typically set as maximizing the accuracy of reconstructing the masked graph components. 
Graph masked autoencoders have attracted increasing attention recently, but it is challenging for them to produce generalizable representations, due to the limitation in data augmentation and reconstruction objective design. To tackle these issues, we propose a more comprehensive graph augmenter in Section~\ref{mask_graph_encode} and an enhanced reconstruction strategy in Section~\ref{Student-Teacher}.

\subsubsection{Edge and Feature Masking.}
\label{mask_graph_encode}
Existing graph generative models commonly employ masking over either edges or features of the input graph, while keeping the other modality intact~\cite{hou2022graphmae, salehi2019graph, park2019symmetric}. Masking both modalities together will negatively affect model learning since it may lack sufficient information for graph reconstruction. However, solely masking and reconstructing one modality limits the model's ability to learn from the other modality, which hinders the learning of comprehensive representations. In this work, we mask both edges and features of the original graph during training. This results in an augmented graph $\mathcal{G}' = f^{aug}(\mathcal{G}) = \{\mathcal{V}, \textbf{A}', \textbf{X}' \}$, where $\textbf{A}'$ and $\textbf{X}'$ denote the masked feature matrix and the masked adjacent matrix, respectively. Formally, we design the augmenter as $f^{aug}(\{\mathcal{V}, \textbf{A}, \textbf{X} \}) =  \{ \mathcal{V}, \textbf{A} \odot \textbf{M}^\text{E}, \text{diag}(\textbf{M}^\text{F}) \cdot \textbf{X}\}$, where $\odot$ stands for the Hadamard product. The edge-mask matrix $\textbf{M}^\text{E} \in \left\{0,1\right\}^{|\mathcal{V}|\times|\mathcal{V}|}$ and the feature-mask matrix $\textbf{M}^\text{F} \in \left\{0,1\right\}^{|\mathcal{V}|}$ are randomly generated binary matrices. The perturbation can be controlled by the sparsity of $\textbf{M}^\text{E}$ and $\textbf{M}^\text{F}$. After data augmentation, we divide the nodes $\mathcal{V}$ into four types: 1) nodes without any masks $\mathcal{V}^N$; 2) nodes with masked edges $\mathcal{V}^E$; 3) nodes with masked features $\mathcal{V}^F$; and 4) nodes with both masked features and edges $\mathcal{V}^B$, where $\mathcal{V} = \mathcal{V}^N \cup \mathcal{V}^E \cup \mathcal{V}^F \cup \mathcal{V}^B$. The masked graph is fed into the encoder to obtain the latent representations $\textbf{Z} = f^{enc}(\textbf{A}', \textbf{X}')$.

\subsubsection{Reconstruction Targets.}
\label{Student-Teacher}
Different from conventional methods that reconstruct edges or features, we propose to reconstruct the embeddings of other graph models as targets. 
Here we consider multiple graph models, and $\textbf{Z}_n \in \mathbb{R}^{|\mathcal{V}|\times d_n}$ denotes the $n$-th reconstruction target.
Different graph models focus on learning different graph information. For example, node2vec~\cite{grover2016node2vec} prioritizes the learning of graph structure information, while PCA~\cite{candes2011robust} trained on feature matrix $\mathbf{X}$ mainly encodes node attribute information.
The information from various modalities is preserved in continuous embedding spaces $\{\textbf{Z}_n\}$ that are homogeneous. By reconstructing these embeddings, we could learn representations that contain both graph topology and attribute information.

Before reconstruction, we further apply re-masking~\cite{hou2022graphmae} on $\textbf{Z}$ to obtain more compressed representations $\textbf{S}\in \mathbb{R}^{|\mathcal{V}|\times 
d}$. Formally, let $\mathbf{s}_i=\textbf{S}[i,:]$ denote the representation of node $v_i$:
\begin{equation}
\mathbf{s}_i = \begin{cases}
  f^{enc}(\textbf{A}', \textbf{X}')[i,:]& \text{ if } v_i \in \mathcal{V}^E \cup \mathcal{V}^F \cup \mathcal{V}^B \\
  \textbf{0}& \text{ if } v_i \in \mathcal{V}^N
\end{cases} ,
\end{equation}
where $\textbf{0}$ is a $d$-dimension all-zero vector. Meanwhile, we also apply re-masking on $\mathbf{Z}_n$ to obtain $\mathbf{T}_n$. Let $\mathbf{t}^n_i=\textbf{T}_n[i,:]$, then $\mathbf{t}^n_i = \textbf{Z}_n[i,:]$ if $v_i \notin \mathcal{V}^N$, and $\mathbf{t}^n_i = \textbf{0} \text{ if} \,\, v_i \in \mathcal{V}^N$.
We implement the decoder with a set of projectors $\{f^{proj}_n(\cdot)\}$, which map $\textbf{S}$ to the target embedding space $\{\mathbf{T}_n\}$. The projected representation is defined as $\textbf{S}_{n} = f^{proj}_{n}(\textbf{S}) \in \mathbb{R}^{|\mathcal{V}|\times d_n}$. Each projector is implemented as a multi-layer perceptron (MLP).

\subsubsection{The Training Objective}
\label{workflow}
 Given the compressed representations $\mathbf{S}$ and target embeddings $\mathbf{T}_1, \mathbf{T}_2,...,\mathbf{T}_n$, the learning objective of our proposed framework is defined as follows:
\begin{equation}
\label{loss}
   \underset{\theta}{\text{min}}   \;\; \mathcal{-L}(\mathbf{S},\{\mathbf{T}_1, \mathbf{T}_2,..., \mathbf{T}_n\}),
\end{equation}
where $\theta$ denotes the trainable parameters in our model, including the parameters in $f^{enc}$ and $\{f^{proj}_n\}$. The objective encourages $\mathbf{S}$ to preserve the knowledge across different targets, which requires an effective and flexible loss function. Previous research has explored various loss functions, such as the $l_2$-norm loss~\cite{salehi2019graph, park2019symmetric} and the cross-entropy loss~\cite{kipf2016variational}. However, the $l_2$-norm loss considers each feature channel independently and neglects their dependencies, leading to sub-optimal results~\cite{tian2019contrastive}. On the other hand, the cross-entropy loss is only applicable to discrete labels. Hence, none of them is suitable for our problem. 
In the next section, we design a new loss to tackle the challenge.

\subsection{Reconstruction Loss Design}
\label{re_loss}

We propose a mutual information (MI) based reconstruction loss to effectively learn knowledge from multiple collaborative targets. To facilitate illustration, we first consider the case of a single target and then extend it to handle multiple targets.

\subsubsection{Single-Target Reconstruction.}
\label{single_teacher}
We start by discussing the scenario where we have a single reconstruction target $\mathbf{T}$. Our goal is to optimize the generative model's representations to capture as much useful information as possible from the target. 
To achieve this, we define our learning objective as maximizing the MI between $\mathbf{S}$ and $\mathbf{T}$, denoted as $\mathcal{L}_\text{Single} = I(\mathbf{S}; \mathbf{T})$. According to Equation~(\ref{infornce2}), the objective can be optimized by maximizing the InfoNCE loss~\cite{oord2018representation}:
\begin{equation}
\label{loss_single}
    \hat{\mathcal{L}}_\text{Single} =  \mathcal{L}_{\mathcal{D}}(\mathbf{S},\mathbf{T}),
\end{equation}
where  $\hat{\mathcal{L}}_\text{Single}$ is a lower bound of $I(\mathbf{S}; \mathbf{T})$. Specifically, the discriminator $\mathcal{D}$ used in $\mathcal{L}_{\mathcal{D}}$ is formulated following~\cite{velivckovic2018deep,tian2020contrastive} as:
\begin{equation}
\label{disc}
    \mathcal{D}(\mathbf{s}, \mathbf{t})=\exp \left(\frac{f^{proj}\left(\mathbf{s}\right) \cdot \mathbf{t}}{\left\|f^{proj}\left(\mathbf{s}\right)\right\| \cdot\left\|\mathbf{t}\right\|} \cdot \frac{1}{\tau}\right),
\end{equation}
where $f^{proj}$ is the projector function, and $\tau$ denotes the temperature hyper-parameter. Intuitively, the information in $\mathbf{t}$ is reconstructed from $\mathbf{s}$ via maximizing the similarity between $\mathbf{t}$ and $f^{proj}(\mathbf{s})$.

\subsubsection{Dual-Target Reconstruction.}
\label{dual_teacher}
We then discuss how to extend the single-target scenario to deal with dual targets that could cover heterogeneous modalities.
A naive design is to maximize the similarity between the projected representation $\mathbf{S}_{n}$ and the target representation $\mathbf{T}_{n}$. 
However, this approach learns from each target individually, failing to capture the information shared between them, which could contain crucial common knowledge that the model should emphasize. To overcome this issue, we propose to quantify this shared knowledge using MI. Specifically, we use $I(\mathbf{S}; \mathbf{T}_1; \mathbf{T}_2)$ to define the \textbf{common knowledge} shared between $\mathbf{T}_1$ and $\mathbf{T}_2$ learned by $\mathbf{S}$. Meanwhile, we use $I(\mathbf{S}; \mathbf{T}_1|\mathbf{T}_2)$ and $I(\mathbf{S}; \mathbf{T}_2|\mathbf{T}_1)$ to define the unique knowledge solely from $\mathbf{T}_1$ and $\mathbf{T}_2$, respectively, that is learned by $\mathbf{S}$. As the two targets could preserve different knowledge, their weights may differ. Hence, we propose to treat each part of knowledge separately by presenting a more general form of Equation~(\ref{loss_single}) as below:
\begin{equation}
\label{loss_mutli}
    \mathcal{L}_\text{Dual} = \lambda_1 I(\mathbf{S}; \mathbf{T}_1|\mathbf{T}_2) + \lambda_2 I(\mathbf{S}; \mathbf{T}_2|\mathbf{T}_1) + \lambda_3 I(\mathbf{S}; \mathbf{T}_1; \mathbf{T}_2).
\end{equation}
The parameters $\lambda_1$, $\lambda_2$, and $\lambda_3$ control the influence of each part of knowledge in model training. A larger value of $\lambda$ indicates a greater importance assigned to the corresponding knowledge. 
In Equation~(\ref{loss_mutli}), it is challenging to directly estimate the conditional mutual information or the multivariate mutual information involving three variables with existing methods~\cite{poole2019variational}.
To address this, we employ chain rules to transform the loss function as:
\begin{align}
     \mathcal{L}_\text{Dual} = &\,\, \lambda_1 [I(\mathbf{S}; \mathbf{T}_1, \mathbf{T}_2) - I(\mathbf{S} ; \mathbf{T}_2)] + \lambda_2 [I(\mathbf{S}; \mathbf{T}_1, \mathbf{T}_2) - I(\mathbf{S}; \mathbf{T}_1)] \notag \\
     & + \lambda_3 [I(\mathbf{S}; \mathbf{T}_1) + I(\mathbf{S}; \mathbf{T}_2) - I(\mathbf{S}; \mathbf{T}_1, \mathbf{T}_2)] \notag  \\
     = & \,\, (\lambda_3-\lambda_2)\cdot I(\mathbf{S}; \mathbf{T}_1) +  (\lambda_3-\lambda_1)\cdot  I(\mathbf{S}; \mathbf{T}_2) \notag  \\
     & + (\lambda_1+\lambda_2-\lambda_3)\cdot I(\mathbf{S}; \mathbf{T}_1, \mathbf{T}_2) \label{loss_mutli2},
\end{align}
where the transformed mutual information can be estimated by the InfoNCE loss. Thus, the dual-target reconstruction loss used for model training is defined as:
\begin{equation}
\label{loss_mutli3}
\begin{split}
     \hat{\mathcal{L}}_\text{Dual} & = \tilde{\lambda}_1 \mathcal{L}_{\mathcal{D}_1}(\mathbf{S}, \mathbf{T}_1) +  \tilde{\lambda}_2 \mathcal{L}_{\mathcal{D}_2}(\mathbf{S}, \mathbf{T}_2) + \tilde{\lambda}_3\mathcal{L}_{\mathcal{D}_3}(\mathbf{S},\{ \mathbf{T}_1, \mathbf{T}_2\}),
\end{split}
\end{equation}
where $\tilde{\lambda}_1 = (\lambda_3-\lambda_2)$, $\tilde{\lambda}_2 = (\lambda_3-\lambda_1)$, $\tilde{\lambda}_3 = (\lambda_1+\lambda_2-\lambda_3)$ are the reorganized weight hyper-parameters. And we set  $\tilde{\lambda}_1, \tilde{\lambda}_2, \tilde{\lambda}_3 \geq 0$.
$\mathcal{L}_{\mathcal{D}_1}$ and $\mathcal{L}_{\mathcal{D}_2}$ apply $\mathcal{D}_1$ and $\mathcal{D}_2$ as their discriminator, which have the same formula as Equation~(\ref{disc}) with $f^{proj}_{1}:\mathbb{R}^{|\mathcal{V}|\times d} \rightarrow \mathbb{R}^{|\mathcal{V}|\times d_1}$ and $f^{proj}_{2}:\mathbb{R}^{|\mathcal{V}|\times d} \rightarrow \mathbb{R}^{|\mathcal{V}|\times d_2}$ as projectors, respectively. $\mathcal{L}_{\mathcal{D}_3}$ applies $\mathcal{D}_3$ as the discriminator which is applicable given three input variables. We define $\mathcal{D}_3$ as:
\begin{equation}
    \mathcal{D}_{3}(\mathbf{s}, \{\mathbf{t}_1,\mathbf{t}_2\})=\exp \left(\frac{f^{proj}_{3}\left(\mathbf{s}\right) \cdot [\mathbf{t}_1;\mathbf{t}_2]}{\left\|f^{proj}_{3}\left(\mathbf{s}\right)\right\| \cdot\left\|[\mathbf{t}_1;\mathbf{t}_2]\right\|} \cdot \frac{1}{\tau}\right),
\end{equation}
where $f^{proj}_{3}:\mathbb{R}^{|\mathcal{V}|\times d} \rightarrow \mathbb{R}^{|\mathcal{V}|\times (d_1+d_2)}$ will map the compressed representations to the concatenated dimension, and $[;]$ denotes concatenation. In particular, when $\lambda_1=\lambda_2$ and $\lambda_3=\lambda_1+\lambda_2$, the knowledge learned from different sources is treated equally.

\subsubsection{Multiple-Target Reconstruction.}
\label{multiple_teacher}
A general version of loss that handles $n\ge3$ targets can be derived from the dual-target loss as:
\begin{equation}
\label{loss_mutli4}
     \hat{\mathcal{L}}_\text{Multi} = \sum_{i \in 2^n-1 } \tilde{\lambda}_{i}\mathcal{L}_{D_{i}}(\mathbf{S}, \{\mathcal{T'}\}_i),
\end{equation}
where $\mathcal{T}=\{\mathbf{T}_1,\mathbf{T}_2,...,\mathbf{T}_n\}$ is the set of target embeddings, and the $\mathcal{T'}=\{\mathbf{T}_1,\mathbf{T}_2,\mathbf{T}_3,...,\{\mathbf{T}_1,\mathbf{T}_2\},\{\mathbf{T}_1,\mathbf{T}_3\},...,\{\mathbf{T}_1,\mathbf{T}_2,...,\mathbf{T}_n\}\}$ is the collective set formed by every subset of $\mathcal{T}$. According to Equation~(\ref{loss_mutli4}), 
as the number of targets increases, the computational cost will increase rapidly. However, in practice, many graph models produce highly similar embeddings containing overlapping information, so we can remove the overlapping targets without affecting the performance. Our experiment results show that state-of-the-art downstream task performance can be achieved with \textbf{no more than three targets}. This observation is also consistent with current research findings on multi-view learning~\cite{tian2020contrastive, hassani2020contrastive}, where too many targets are not necessarily needed for desirable results.

\subsection{Reconstruction Target Candidates}
\label{teacher}
In this subsection, we discuss how to choose reconstruction targets and how to choose their weights.

\subsubsection{Reconstruction Target.}
\label{candidates}
A wide range of graph embedding models are potential candidates. We categorize the candidates into three groups based on their input and inductive bias. 

\paragraph{Target embeddings with structural knowledge.}
This group of target embeddings captures the original graph's topological or structural information. Examples include Grarep~\cite{ahmed2013distributed}, Deepwalk~\cite{perozzi2014deepwalk}, and node2vec~\cite{grover2016node2vec}. In this paper, we use the output of \textbf{node2vec} as our structural target embedding. Given a graph $\mathcal{G} = \{\mathcal{V}, \textbf{A}\}$ as input, node2vec generates target embedding that preserves path-aware topological information by employing two traversing strategies: breadth-first sampling (BFS) and depth-first sampling (DFS). The combination of the two strategies enables node2vec to learn comprehensive structural information. Compared with the recent self-supervised learning models (e.g., contrastive learning), node2vec is more efficient due to its lower time and space complexity.

\paragraph{Target embeddings with attribute knowledge.}
This group of target embeddings encodes attribute information from an input graph. Examples include principal component analysis (PCA)~\cite{candes2011robust} and autoencoder~\cite{wang2016auto}. In this paper, we choose the output of \textbf{PCA} as our attribute target embedding. Given a graph $\mathcal{G} = \{\mathcal{V}, \textbf{X}\}$ as input, PCA maps the feature matrix into a lower-dimensional embedding space, which preserves most of the useful information and eliminates the noise~\cite{salmon2014poisson}. Here PCA is also efficient due to its relatively low time and space complexity. 
In the rest of the paper, \textbf{we choose node2vec and PCA as default models} to provide embeddings as the reconstruction targets, since they are efficient and can encode distinct information modalities.

\paragraph{Target embeddings with hybrid knowledge.}
This group leverages GNNs as encoders to capture both topological and feature information. Examples include graph contrastive learning models~\cite{zhu2020deep, you2020graph, thakoor2021bootstrapped} and graph generative models~\cite{kipf2016variational, salehi2019graph}. In this paper, we select the Graph Autoencoder (GAE) as the hybrid knowledge extractor~\cite{kipf2016variational}. Given a graph $\mathcal{G} = \{\mathcal{V},\textbf{A}, \textbf{X}\}$ as input, the GAE learns target embeddings by reconstructing edges of the original graph. 
However, GAE comes with a higher computational cost, thus we only include it for performance comparison in the ablation study.

\subsubsection{Weight Setting Strategy.}
\label{weight}
In this paper, we decide the weight value of $\{\lambda_1, \lambda_2, ...\}$ based on the masking results of $f^{aug}$. 
The key idea is to prioritize the learning of information that is masked by $f^{aug}$, as shown in Figure~\ref{design}.
For example, suppose we use node2vec as $target_1$ and PCA as $target_2$. For nodes in $\mathcal{V}^E$ whose edges are masked, since $target_1$ contains the topological information, we prioritize the learning of this information, meaning that $\lambda_1 > \lambda_2$.
In contrast, for nodes in $\mathcal{V}^F$, a reasonable setting would be $\lambda_2>\lambda_1$. 
In self-supervised learning, this would make the pre-training task more challenging and valuable, encouraging the encoder to learn more generalizable patterns from the input~\cite{he2022masked,hou2022graphmae}.  

Furthermore, $\mathcal{V}^B$ consists of nodes with masked edges and features, which results in the absence of both structural and attribute information, making their reconstruction particularly challenging and unstable. To address this, we prioritize the reconstruction of the most fundamental information for nodes in $\mathcal{V}^B$. Intuitively, this information refers to the shared knowledge among all target models, which we refer to as common knowledge in Section~\ref{dual_teacher} and~\ref{multiple_teacher}. For example, in the dual-target scenario, a reasonable weight setting for nodes in $\mathcal{V}^B$ would be $\lambda_3>\lambda_1$ and $\lambda_3>\lambda_2$. We validate the effectiveness of our weight setting strategy through an ablation study in our experiments.

\begin{table*}[htb]
  \caption{Node classification performance comparison (Accuracy). A.R. means the average rank.}
  \label{node_classification}
  \centering
  \vspace{-10pt}
\resizebox{1.9\columnwidth}{!}{
\begin{tabular}{cl|ccccccc|>{\columncolor{mygray}}c}
\hline
Model Type &Dataset             & Cora          & Citeseer         &WikiCS  &Computers  &Photo &CS &Physics  &A.R. \\ \hline \hline
Randwalk &Deepwalk       &73.32±0.61   &48.24±1.35   &76.37±0.19 &86.59±0.10  &90.08±0.29 &85.40±0.13 &92.37±0.08 &8.86   \\\hline
\multirow{3}{*}{Contrastive} &MVGRL           &84.39±0.34 &71.71±0.24 &80.15±0.27 &88.28±0.13 &92.29±0.19 &\textbf{92.91 ±0.07} &95.36±0.06  &  4.71
 \\ 
&GCA             &83.86±0.45     &71.83±0.68     &78.86±0.26 &88.06±0.12 &92.44±0.29 &92.66±0.09 &95.50±0.18 &5.71    \\
&BGRL        &82.35±0.26     &71.29±0.59     &79.47±0.26  &89.80±0.17   &92.77±0.15  &92.60±0.05  &95.60±0.04 &4.86 \\\hline
\multirow{4}{*}{Generative}&VGAE            &80.62±0.32  &71.85±1.19   &78.21±0.43 &81.60±0.28  &90.77±0.44 &92.41±0.13 &95.24±0.03 &7.43    \\
&GraphMAE        &84.23±0.42     &\textbf{72.74±0.29}     &80.57±0.17  &89.39±0.84  &92.83±0.43  &92.68±0.30 &95.52±0.15   &3.29\\ 
&GraphMAE2        &84.41±0.30      &72.11±0.42     &81.01±0.34   &89.50±0.76   &92.87±0.14 &92.74±0.14  &95.38±0.08    &3.00  \\ 
&S2GAE       &84.14±0.65      &72.21±0.47      &79.14±0.23  &89.64±0.12   &92.09±0.28   &89.92±0.17  &94.37±0.14    & 5.71 \\ 
\hline\hline
Generative &GiGaMAE   &\textbf{84.72±0.47} &72.31±0.50 &\textbf{81.14±0.16}   &\textbf{90.45±0.16}   &\textbf{93.01±0.41}   &92.72±0.32 &\textbf{95.66±0.14} &\textbf{1.43}\\
\hline
\end{tabular}}
\vspace{-5pt}
\end{table*}

\begin{table*}[htb]
  \caption{Node clustering performance comparison (NMI/ARI).}
  \label{node_clustering}
  \centering
\vspace{-10pt}
\resizebox{1.90\columnwidth}{!}{
\begin{tabular}{l|ccccccc|>{\columncolor{mygray}}c}
\hline
Dataset              & Cora          & Citeseer         &WikiCS  &Computers  &Photo  &CS &Physics &A.R. \\ \hline
\hline
Deepwalk       &0.4161/0.3416   &0.1376/0.1452    &0.4660/0.3587  &0.4202/0.2637   &0.6482/0.5129  &0.6445/0.4863  &0.6995/0.7985 &5.21\\
MVGRL           &0.5481/0.5167     &0.4073/0.4115    &0.2135/0.1101  &0.2657/0.1806   &0.1776/0.1127  &0.6436/0.4737  &0.4948/0.4799 &6.79 \\ 
GCA          &0.4645/0.3268     &0.2681/0.2178     & 0.1463/0.0176  &0.4062/0.1512   &0.4480/0.2518  &0.6975/0.5578  &0.6638/0.7468  &6.93\\
BGRL        &0.2851/0.0920     &0.2156/0.1759     &0.2767/0.0937  &0.4396/0.2096   &0.6189/0.4754  &\textbf{0.7740}/0.6422  &0.7249/0.8130 &5.50\\
VGAE            &0.4930/0.4392 &0.4104/0.4247 &0.3453/0.1478 &0.3073/0.2054 &0.4847/0.3539  &0.7736/\textbf{0.6646}  &0.4925/0.2628  &5.50\\
GraphMAE       &0.5781/0.5082     &\textbf{0.4330/0.4423}     &0.4038/0.2951  &0.5015/0.3298   &0.6676/0.5703  &0.7297/0.5691  &0.6348/0.6734 &3.14\\ 
GraphMAE2        &0.5821/0.5310      &0.4283/0.4268      &0.3674/0.2541   &0.5053/0.3418   &0.6496/0.5613   &0.4423/0.2449  &0.2820/0.1564    &4.50 \\ 
S2GAE       &0.5127/0.4481      &0.3346/0.2830      &0.3143/0.1110   &0.4397/0.2297   &0.5624/0.3427   &0.6251/0.4289  &0.6152/0.7059    &5.93 \\ 
\hline\hline
GiGaMAE   &\textbf{0.5836/0.5453} &0.4224/0.4283 &\textbf{0.4910/0.4239}   &\textbf{0.5228/0.3579}   &\textbf{0.7066/0.5859}   &0.7622/0.6417 &\textbf{0.7373/0.8271} &\textbf{1.50} \\ 
\hline
\end{tabular}}
\vspace{-5pt}
\end{table*}

\begin{table*}[htb]
  \caption{Link prediction performance comparison.}
  \label{link_prediction}
  \centering
  \vspace{-10pt}
\resizebox{1.90\columnwidth}{!}{
\begin{tabular}{lc|ccccccc|>{\columncolor{mygray}}c}
\hline
Dataset          &Metrics    & Cora          & Citeseer         &WikiCS  &Computers  &Photo &CS &Physics &A.R. \\ \hline
\hline

\multirow{2}{*}{Deepwalk}  &AUC  &76.33±0.48  &64.67±0.61      &91.06±0.06      &87.36±0.06   &91.74±0.09 &91.01±0.12 &91.94±0.07 &   \\
 &AP &81.77±0.29       &72.77±0.49      &91.76±0.08      &87.34±0.04   &91.55±0.08 &92.28±0.10 &91.93±0.07   &\multirow{-2}{*}{6.14}\\
\multirow{2}{*}{MVGRL} &AUC           &74.57±0.38     &68.33±0.59      &93.09±0.14   &85.32±0.25 &84.89±0.08 &77.13±0.33 &77.26±0.53 &  \\
&AP &77.16±0.28      &72.79±0.32      &93.37±0.15   &86.45±0.21 &85.54±0.10 &78.77±0.35 &79.84±0.42 &\multirow{-2}{*}{7.64} \\
\multirow{2}{*}{GCA}  &AUC  &89.88±1.21      &87.25±0.87      &93.83±0.23   &90.95±0.44 &91.47±0.47 &87.72±0.21 &90.45±0.34  &  \\
&AP &89.39±1.76     &87.59±0.65      &93.90±0.35   &89.80±0.13 &91.26±0.46 &86.08±0.34 &88.12±0.37  &\multirow{-2}{*}{5.71}  \\
\multirow{2}{*}{BGRL} &AUC         &91.70±0.59      &92.90±0.57      &89.67±0.58   &93.69±0.43 &94.21±0.64 &92.60±0.15 &92.29±0.56 &  \\
&AP&92.18±0.43       &93.91±0.44      &89.67±0.59   &92.86±0.53 &93.44±0.65 &91.29±0.19 &90.79±0.88 & \multirow{-2}{*}{4.36}\\
\multirow{2}{*}{VGAE}  &AUC    &94.18±0.80       &93.79±0.21      &96.99±0.11      &94.09±0.10  &95.28±0.14 &95.94±0.14 &95.88±0.17 & \\
&AP &95.12±0.11        &93.80±0.23      &97.75±0.46      &93.84±0.11   &94.59±0.16 &95.41±0.17 &95.24±0.10 &\multirow{-2}{*}{2.36} \\
\multirow{2}{*}{GraphMAE} &AUC         &88.78±0.87      &90.32±1.26      &71.40±4.19   &74.04±3.08 &74.58±3.90 &85.37±1.37 &81.29±5.13 &  \\
&AP&88.32±0.91       &91.54±0.87      &68.60±3.96  &70.08±2.80  &72.30±3.01 &83.93±1.08 &79.71±4.40 &\multirow{-2}{*}{7.86}  \\
\multirow{2}{*}{GraphMAE2} &AUC         &89.54±0.30       &90.48±0.98    &72.71±3.33  &73.99±3.04  &83.77±1.32   &89.22±0.35  &83.20±1.43 &  \\
&AP&88.91±0.31       &91.53±0.71       &69.66±3.31    &70.05±2.76  &80.71±1.08  &87.13±0.32   &82.24±1.05  &\multirow{-2}{*}{7.07}  \\
\multirow{2}{*}{S2GAE} &AUC         &93.12±0.58       &93.81±0.23       &\textbf{98.74±0.02}    &94.59±1.16  & 93.84±2.22  &96.13±0.48   &95.21±0.75  &  \\
&AP&93.96±0.58       &94.21±0.24       &\textbf{98.76±0.02}    &\textbf{94.01±1.11}  &92.07±3.12  &\textbf{95.73±0.62}   &94.56±0.83  & \multirow{-2}{*}{2.29} \\
\hline\hline
\multirow{2}{*}{GiGaMAE}  &AUC    &  \textbf{95.13±0.15} &\textbf{94.18±0.36}      &95.30±0.09      &\textbf{95.17±0.38}   &\textbf{96.24±0.11} &\textbf{96.34±0.07} &\textbf{96.32±0.08} & \\
&AP &\textbf{95.20±0.17}   &\textbf{94.40±0.12}     &95.59±0.09      &92.91±0.51   &\textbf{94.62±0.11} &95.28±0.09 &\textbf{95.27±0.04} &\multirow{-2}{*}{\textbf{1.57}}  \\
\hline
\end{tabular}}
\vspace{-5pt}
\end{table*}

\section{Experiments}
We conduct experiments to answer the following questions. \textbf{Q1}: Does the proposed method generalize well to the common downstream tasks?  \textbf{Q2}: How does our framework perform given different targets with different settings?  \textbf{Q3}: Is our reconstruction loss effective with the proposed masking and weight setting strategy?  

\subsection{Experiments Setup}

\subsubsection{Datasets and Baselines} We demonstrate the effectiveness of our framework on various node-level tasks. We select seven representative benchmark datasets~\cite{yang2016revisiting,mernyei2020wiki,shchur2018pitfalls}, including Cora, CiteSeer, WikiCS, Amazon-Computers (Computers), Amazon-Photo (Photo), Coauthor-CS (CS), and Coauthor-Phy (Phy), as benchmarks. Our proposed framework is compared with various types of state-of-the-art models. The first type is contrastive learning models, which include GCA~\cite{zhu2021graph}, MVGRL~\cite{hassani2020contrastive}, and BGRL~\cite{thakoor2021bootstrapped}. Among them, GCA uses node centrality to generate high-quality contrastive views; MVGRL utilizes multiple views of graphs for contrastive learning; BGRL gets rid of negative examples by leveraging the idea of self-distillation. The second type is generative models, which include VGAE~\cite{kipf2016variational}, GraphMAE~\cite{hou2022graphmae}, GraphMAE2~\cite{hou2023graphmae2}, and S2GAE~\cite{tan2023s2gae}. The VGAE and S2GAE aim to predict the existence of edges, and GraphMAE and GraphMAE2 seek to reconstruct the node feature. The last type is a random-walk based model, i.e., Deepwalk~\cite{perozzi2014deepwalk}. 

\subsubsection{Experimental Settings}

Our proposed framework is implemented in PyTorch~\cite{paszke2019pytorch} and PyG (PyTorch Geometric)~\cite{Fey/Lenssen/2019}. The implementation details for baselines and GiGaMAE can be found in the Appendix. Unless otherwise specified, we obtain our target embeddings $\mathbf{T}_1$ and $\mathbf{T}_2$ using \textbf{node2vec} and \textbf{PCA}, respectively.

\begin{figure}[h]
  \begin{adjustwidth}{-0.5cm}{}
    \centering
    \includegraphics[width=0.40\textwidth]{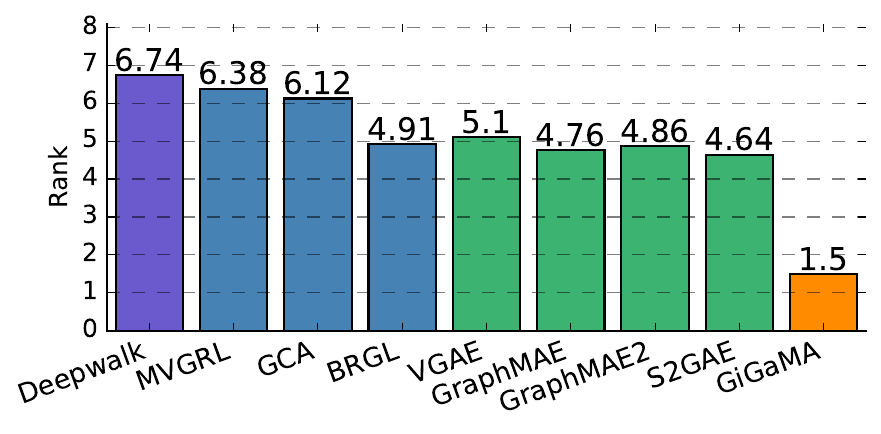}
\end{adjustwidth}
    \vspace{-10pt}
    \caption{The average of A.R. on three tasks (A lower value indicates better performance and generalization ability).}
    \label{fig:ar_avg}
    \vspace{-5pt} 
\end{figure}

\subsection{Quantitative Evaluation} 
To answer \textbf{Q1}, we benchmark the performance of GiGaMAE on three crucial graph learning tasks: node classification, node clustering, and link prediction. For a clear comparison, we report the average rank (A.R.) along with the corresponding metric scores. We also provide the average value of A.R. on these tasks in Figure~\ref{fig:ar_avg}. We can observe that \textbf{GiGaMAE is the only method that consistently performs well over the three tasks}, which validates its generalization ability. Detailed analysis is as follows.

\subsubsection{Node Classification} 
\label{classificaiton}
For the node classification task, we follow the same linear evaluation protocol in~\cite{zhu2020deep}. First, we use the pre-trained model to generate node representations given the whole graph. Then we evaluate the node representation quality under a logistic-regression classifier using grid-search, where the train/test split follows~\cite{zhu2020deep, zhu2021graph}. It is important to point out that the applied split setting is different from the ones used in~\cite{thakoor2021bootstrapped, hassani2020contrastive, tan2023s2gae, hou2022graphmae,hou2023graphmae2}, resulting in slight variations in the reported results. Finally, we report the mean accuracy results with standard deviation on the test nodes for ten runs. 
The results are listed in Table~\ref{node_classification}. Our proposed model achieves the best performance in most datasets and ranks top-1 over eight state-of-the-art baselines.

\subsubsection{Node Clustering}
\label{clustering}
For the node clustering task, we follow the evaluation protocol in~\cite{park2019symmetric, wang2017mgae}.  The whole data is used in the pre-training process to obtain representations, which are then clustered using the K-Means algorithm with cluster numbers set to label class numbers. We report the mean value of normalized mutual information (NMI) and average rand index (ARI) for ten runs.
Table~\ref{node_clustering} reports the clustering results. Our approach has the best overall clustering performance ranking top-1 against the baselines. We also find that the models (e.g., MVGRL) with good classification performance do not always have good clustering performance, suggesting these two tasks require different kinds of knowledge.

\begin{table*}[htb]
  \caption{Target embeddings performance vs. GiGaMAE performance.}
  \label{transfer ability}
  \centering
  \vspace{-10pt}
\resizebox{1.90\columnwidth}{!}{
\begin{tabular}{l|ccc|ccc|ccc}
\hline
Dataset              & \multicolumn{3}{|c|}{Cora} &\multicolumn{3}{|c|}{WikiCS} &\multicolumn{3}{|c}{Computers}  \\
Metrics      &ACC &NMI/ARI  &AUC  &ACC &NMI/ARI  &AUC  &ACC &NMI/ARI  &AUC   \\
\hline
\hline
Node2vec     &71.76±0.75 &0.3944/0.2460  &85.85±1.13  &71.57±0.36 &0.4081/0.3452 &91.70±0.26  &83.98±0.28  &0.5248/0.3882  &84.83±0.21 \\
PCA       &42.22±0.57 &0.0212/0.0068  &68.93±0.05  &68.52±0.25  &0.3181/0.2211 &79.03±0.02 &66.40±0.25  &0.0305/0.0236  &59.86±0.02 \\
GAE          &81.29±0.61 &0.4907/0.4136  &90.04±0.15  &70.33±0.86  &0.1091/0.0603 &94.65±0.16 & 77.01±1.08  &0.3204/0.1749  &91.58±0.13 \\
\hline\hline
GiGaMAE$_{Node2vec}$    &84.00±0.51 &0.5376/0.4706  &93.67±0.19  &80.56±0.51  &0.4687/0.3592 &94.60±0.44 &89.98±0.21  &0.5089/0.3207  &87.71±0.64  \\
GiGaMAE$_{PCA}$  &83.17±0.59 &0.5343/0.4861  &93.44±0.60  &80.96±0.33 &0.4810/0.3397 &94.59±0.21 &90.07±0.16 &0.5139/0.3123  &89.60±1.97  \\
GiGaMAE$_{GAE}$    &84.14±0.63 &0.5802/0.5257  &92.78±0.31  &80.74±0.30 &0.4664/0.3433   &94.16±0.13 &89.46±0.06  &0.5288/0.3843  &87.58±0.08  \\ \hline \hline
GiGaMAE    &84.72±0.47 &0.5836/0.5453  &95.13±0.15  &81.14±0.16 &0.4908/0.4211 &95.30±0.09 &90.45±0.16  &0.5228/0.3579  &95.17±0.38  \\
GiGaMAE$_{Large}$    &84.69±0.46  &0.5808/0.5394   &95.12±0.09   &81.14±0.20  &0.4777/0.3748  &95.34±0.04  &90.44±0.20   &0.5375/0.3901   &93.61±0.29   \\
\hline
\end{tabular}}
\vspace{-5pt}
\end{table*}

\begin{table*}[htb]
  \caption{Models performance with naive learning objectives and mask strategies.}
  \label{learning and mask}
  \centering
  \vspace{-10pt}
\resizebox{1.90\columnwidth}{!}{
\begin{tabular}{l|l|ccc|ccc|ccc}
\hline
Dataset        &      & \multicolumn{3}{|c|}{Cora} &\multicolumn{3}{|c|}{WikiCS} &\multicolumn{3}{|c}{Computers}  \\ 
Metrics  &    &ACC &NMI/ARI  &AUC  &ACC &NMI/ARI  &AUC &ACC &NMI/ARI  &AUC   \\
\hline
\hline
\multirow{3}{*}{Naive Integration}& MaxPooling  &47.24±0.56 &0.0621/0.0308  &64.05±0.02  &57.67±0.38  &0.1070/0.0536 &60.27±0.02 &65.36±0.27  &0.0569/0.0067  & 44.44±0.03  \\
&AvgPooling &58.16±0.88 &0.3117/0.2537  &72.92±0.03  &68.88±0.27  &0.1462/0.0661 &86.15±0.01 &74.20±0.30  &0.3414/0.2133  &74.87±0.02   \\
&Concatenate &70.94±0.48 &0.4119/0.3641  &77.59±0.01  &74.51±0.30  &0.1625/0.0767 &89.84±0.01 &82.71±0.19  &0.3715/0.2218  &81.90±0.02   \\
\hline\hline
\multirow{3}{*}{Naive Loss} &MSE  &83.83±0.70 &0.5862/0.5425  &56.68±0.13  &80.43±0.39  &0.4623/0.3639 &89.32±0.02 &89.85±0.76  &0.5182/0.3469  &72.75±0.68  \\
&Scaled Cosine  &84.10±1.18 &0.5489/0.4722  &94.22±0.35  &80.95±0.20  &0.4586/0.3456 &91.04±0.02 &89.82±0.23  &0.5128/0.3254  &89.36±1.01   \\
&Contrastive Loss &84.17±0.46 &0.5792/0.5363  &94.61±0.10  &80.76±0.25  &0.4870/0.4163 &92.58±0.60 &90.03±0.30  &0.5080/0.3045  &90.63±1.75   \\\hline\hline
\multirow{2}{*}{Naive Mask}  
&w/o Mask Edge &84.22±0.30  &0.5706/0.5329   &94.01±0.10   &80.91±0.15   &0.4912/0.4194  &91.90±0.08  &90.25±0.13   &0.5056/0.3005   &93.97±0.26    \\
&w/o Mask Feature &83.58±0.34  &0.5683/0.5235   &94.94±0.27   &80.24±0.15   &0.4846/0.4087  &94.10±0.25  &90.03±0.14   &0.5279/0.3613   &94.29±0.52   \\ \hline\hline
GiGaMAE  &   &84.72±0.47 &0.5836/0.5453  &95.13±0.15  &81.14±0.16 &0.4908/0.4211 &95.30±0.09 &90.45±0.16  &0.5228/0.3579  &95.17±0.38  \\
\hline
\end{tabular}}
\vspace{-5pt}
\end{table*}

\begin{figure*}[h]
    \begin{subfigure} [t]{0.255\textwidth}
    \includegraphics[width=0.98\textwidth]{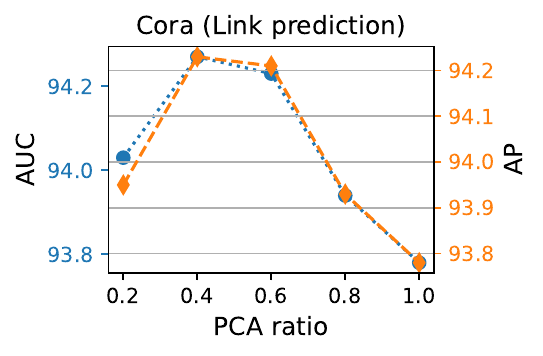}
    \end{subfigure}
    \hspace{-13pt}
    \begin{subfigure} [t]{0.255\textwidth}
    \includegraphics[width=0.956\textwidth]{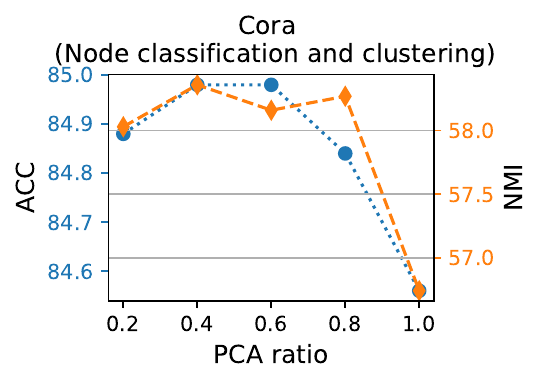}
    \end{subfigure}
    \hspace{-14pt}
    \begin{subfigure} [t]{0.26\textwidth}
    \includegraphics[width=1.01\textwidth]{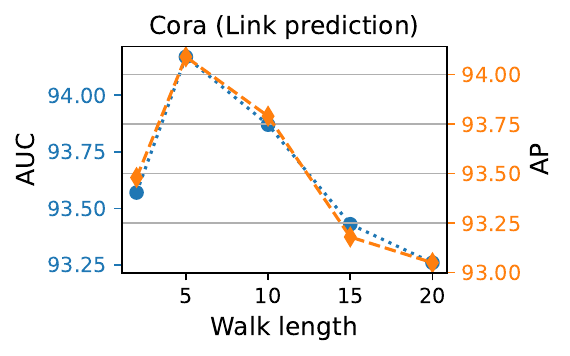}
    \end{subfigure}
    \hspace{-10pt}
    \begin{subfigure} [t]{0.255\textwidth}
    \includegraphics[width=0.940\textwidth]{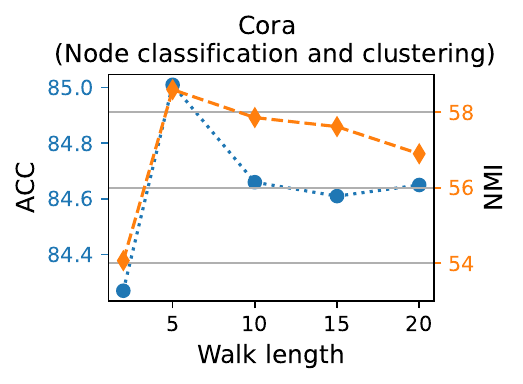}
    \end{subfigure}
    \vspace{-10pt}
    \caption{Downstream task performance with different PCA ratios and node2vec walk length on Cora.} \label{pca}
    \vspace{-8pt}
\end{figure*}

\begin{figure}[h]
    \begin{subfigure}[t]{0.22\textwidth}
    \includegraphics[width=1\textwidth]{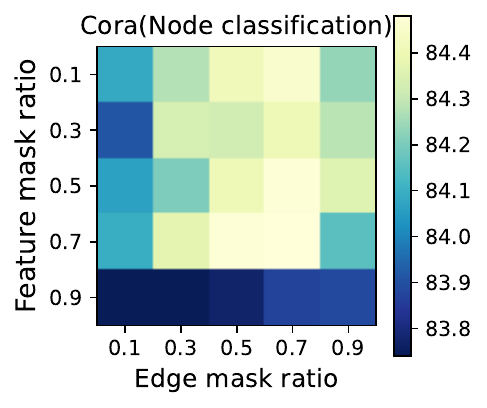}
    \end{subfigure}
    \hspace{-1pt}
    \begin{subfigure}[t]{0.22\textwidth}

    \includegraphics[width=0.95\textwidth]{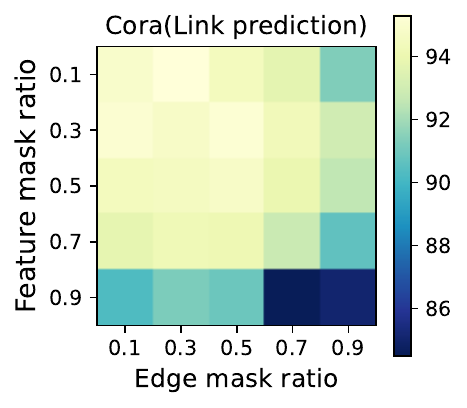}
    \end{subfigure}
    \vspace{-3mm}
    \caption{Downstream task performance with different feature/edge mask ratios on Cora.} \label{mask_ratio}
    \vspace{-3mm}
\end{figure}

\begin{figure}[h]
    \begin{subfigure} [t]{0.246\textwidth}
    \includegraphics[width=0.96\textwidth]{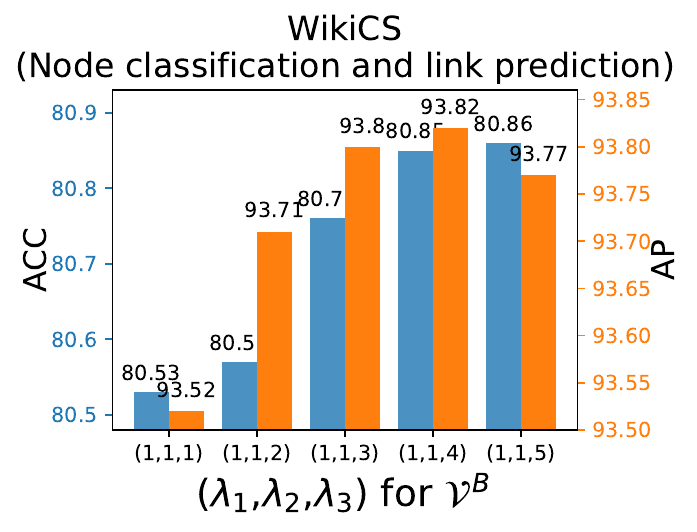}
    \end{subfigure}
    \hspace{-14pt}
    \begin{subfigure} [t]{0.246\textwidth}
    \includegraphics[width=0.95\textwidth]{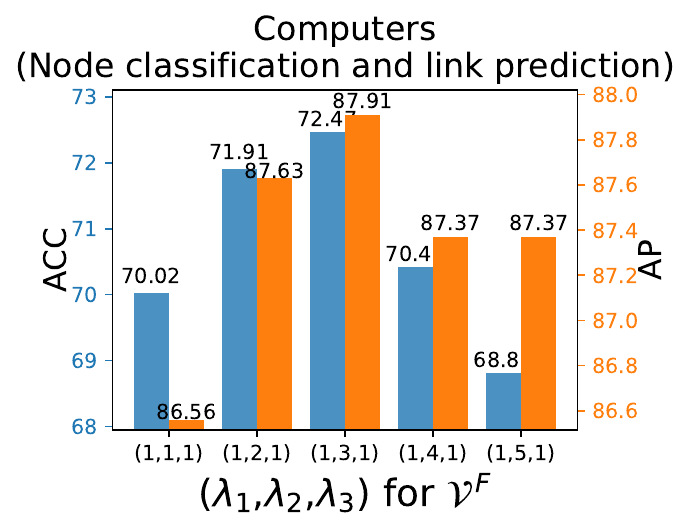}
    \end{subfigure}
    \vspace{-10pt}
    \caption{Downstream task performance with different weight settings on WikiCS and Computers.} \label{weight_setting}
    \vspace{-12pt}
\end{figure}

\subsubsection{Link Prediction}
\label{prediction}
For the link prediction task, we use the same evaluation protocol as in~\cite{tan2023s2gae, tan2023bring, tan2023collaborative}. The whole dataset is split into three parts: the training set (85\%), the validation set (5\%), and the test set (10\%). We only use the training set to train the GiGaMAE model and the validation set to tune the hyperparameters. The reported AUC and Average Precision (AP) scores are calculated on the test set. The link prediction results are shown in Table~\ref{link_prediction}. We make three observations: (1) Among baseline methods, S2GAE and VGAE show the best results, which makes sense since their pre-training task is to infer node connections. (2) The contrastive learning methods (GCA, MVGRL, BGRL) have the second-best performance. The edge data augmentation in contrastive learning enables these models to pick up structural information. (3) GraphMAE and GraphMAE2 show the worst performance on the link prediction task, where a possible reason is that they focus on learning attribute information since they mainly reconstruct features of graphs.

\subsection{Ablation Studies on Embedding Models} 
We answer \textbf{Q2} in this section. The experiments show that even with just one target, our obtained representations can still achieve better performance than the target embeddings, demonstrating the effectiveness of our self-supervised learning design. Meanwhile, the performance and generalization ability of our framework could be further improved with multiple targets.
The computational cost analysis is provided in the Appendix.

\subsubsection{Evaluation with Different Targets}
\label{transfer}
Table~\ref{transfer ability} presents the performance of target embeddings and their corresponding GiGaMAE performance. The metric ACC denotes the accuracy of node classification. The metrics used to evaluate node clustering are NMI and ARI, which indicate the mean value of normalized mutual information and average rand index, respectively. The metric AUC evaluates link prediction performances. We report the average values (with standard deviation) after running experiments five times. 

The results verify that our GiGaMAE trained on a single target outperforms the target embeddings. This finding shows that the mask-and-reconstruction paradigm empowers the learned representations with informative knowledge that is beyond the original reconstruction target. For the node classification task, our proposed GiGaMAE model can improve the accuracy by 14.43\%. However, learning from a sole target cannot ensure the GiGaMAE model maps the original node into generalizable representations. Therefore, we need to introduce multiple targets. We report the performance of GiGaMAE (with node2vec and PCA as embedding models) and GiGaMAE$_{Large}$ (with node2vec, PCA, and GAE as embedding models). 
Both of them demonstrate satisfying performance.

\subsubsection{Evaluation on Embedding Models Settings}
\label{teacher_setting}
Figure~\ref{pca} demonstrates the influence of embedding models' hyper-parameter settings on the proposed framework. Specifically, we choose the compression ratio of PCA and the walk length of node2vec for analysis. The experiments are conducted on the Cora dataset. We can observe that a too-small or too-large PCA ratio could impair the downstream task performance. On the one hand, reconstructing the heavily compressed embedding (ratio=0.2) will lead to the degradation of link prediction and node classification/clustering task performance, possibly due to the reconstructed targets lacking sufficient information. On the other hand, reconstructing the full-size feature matrix (ratio=1.0) will also impair downstream task performance, possibly due to information redundancy. A similar trend can also be observed in the walk length setting. A moderate value (length =5) leads to a decent performance, while a too-small (length =2) or too-large (length =20) value will hurt the performance.

\subsection{Ablation Studies on Autoencoder Models} 
We answer \textbf{Q3} in this section, where we validate the effectiveness of our proposed learning and mask strategies along with our proposed reconstruction loss design and weight setting strategy.

\subsubsection{Evaluation on Learning Strategies}
\label{Strategies}
Our GiGaMAE effectively integrates knowledge from multiple targets. For comparison, we design three naive knowledge integration methods, with results in Table~\ref{learning and mask}. In the naive methods, the obtained target embeddings are directly used in downstream tasks after a simple aggregation operation. The aggregation operations include maxpooling, avgpooling, and concatenate. We can observe that these naive methods have poor performance indicating that they are not valid knowledge integration methods compared with the graph masked autoencoder approaches. The impact of different learning objectives is also analyzed in Table~\ref{learning and mask}.
For comparison, we replace our proposed MI-based reconstruction loss with MSE, scaled cosine~\cite{hou2022graphmae} and multi-view contrastive loss~\cite{tian2020contrastive}, and leave everything else unchanged. The result shows that the scaled cosine and contrastive loss perform better than the MSE. However, neither of them can match the effectiveness of our proposed MI-based learning objective.

\subsubsection{Evaluation on Mask Strategies}
\label{mask_strategies}

We report the performance of our proposed framework with only one kind of data augmentation, i.e., w/o edge masking and w/o feature masking. The results in Table~\ref{learning and mask} show that the sole edge or feature mask strategy degrades the downstream performance and generalization ability, which proves the necessity of combining two kinds of mask augmentation together in the generative models. Then we apply different mask ratios on the Cora dataset to see how the downstream performance will change. The result is shown in Figure~\ref{mask_ratio}, which shows that a larger mask ratio benefits the node classification task while a smaller mask ratio is preferred for the link prediction task.

\vspace{-3pt}
\subsubsection{Evaluation on Weight Setting}
\label{knowledge weight setting}
Figure~\ref{weight_setting} demonstrates the effectiveness of our proposed weight setting strategy. In the WikiCS dataset, the weight $\lambda_3$ for the shared knowledge introduced by both node2vec and PCA embeddings is gradually increased for nodes $\mathcal{V}^B$ (mask both edges and features). 
In the Computers dataset, the weight $\lambda_2$ for the knowledge solely introduced by the PCA embedding is gradually increased from 1 to 5 for nodes $\mathcal{V}^F$ (only mask features). To eliminate interference from other types of nodes, we only use $\mathcal{V}^B$ and $\mathcal{V}^F$ to calculate the loss in the above cases, respectively. In the left figure, we observe that the model performance improves as we increase $\lambda_3$, indicating that the shared knowledge between two targets benefits the learning of $\mathcal{V}^B$ nodes, which is consistent with our hypothesis in loss function design. In the right figure, the model performance on $\mathcal{V}^F$ nodes improves as $\lambda_2$ increases, indicating PCA embedding is more important for self-supervised learning when attributes (of its input graph) are masked.

\vspace{-3pt}
\section{Related Work: Graph Autoencoder}
\label{Related Work}

Graph autoencoders typically reconstruct graph components such as edges or features. A traditional way to achieve such reconstruction is to enforce the model to recover the original input graph data. Examples of early research include GAE and VGAE~\cite{kipf2016variational}, which predict link existence, GALA~\cite{park2019symmetric}, which reconstructs features, and GATE~\cite{salehi2019graph}, which reconstructs both edges and features. However, these models suffer from overfitting issues and do not produce robust representations~\cite{vincent2008extracting}. To address these challenges, recent works have adopted self-supervised learning strategies, leveraging data augmentation techniques to encourage the model to learn more informative underlying patterns. For example, S2GAE~\cite{tan2023s2gae} masks edges in the graph and predicts missing links, while MaskGAE~\cite{li2022maskgae} corrupts both edge and path and reconstructs the original edge and degree information. GraphMAE~\cite{hou2022graphmae} utilizes GNN models as the encoder and decoder to reconstruct masked node features, while GraphMAE2~\cite{hou2023graphmae2} introduces latent representation prediction with random re-masking for node attribute reconstruction. Although these models have shown superior performance, they focus on reconstructing specific modality information, limiting their ability to capture more comprehensive knowledge. GPT-GNN~\cite{hu2020gpt} tries to overcome this by incorporating both edge masking and feature masking in the training process, aiming to reconstruct joint probability distributions of the graph. However, it assumes a sequential dependency among the generated features/edges that may not exist in real graphs, which generally limits the model applicability. As a result, an effective generative model that seamlessly combines structural and attribute reconstruction is still lacking. To address this issue, we propose the GiGaMAE framework.

\section{Conclusion}
In this paper, we present a novel framework for Generalizable Graph Masked Autoencoder (GiGaMAE). Our GiGaMAE learns generalizable node representations by reconstructing target embeddings that contain diverse information.  
We also design a new reconstruction loss based on mutual information, which is flexible to handle knowledge learned by targets individually and shared between multiple targets. We evaluate GiGaMAE via extensive experiments using two efficient target models (node2vec and PCA), where GiGaMAE consistently shows good performance on seven benchmark datasets and three graph learning tasks.

\begin{acks}
The work is, in part, supported by NSF (\#IIS-2223768, \#IIS-2223769). The views and conclusions in this paper
are those of the authors and should not be interpreted as representing any funding agencies.
\end{acks}

\appendix

\section{Appendix}

\subsection{Computational Cost Comparison}

Our approach requires training the embedding model before reconstruction, resulting in additional computation costs. In this section, we compare the training time of GiGaMA with other baseline models, as shown in Table~\ref{ComputationCost}. Our model requires more computation time than GraphMAE due to the extra training time for the embedding models and more advanced learning objective. However, our framework is faster than the contrastive model GCA as we only use a partial graph (masked nodes) for each epoch's loss calculation, reducing the computation cost.

\begin{table}[h]
\renewcommand\arraystretch{1.2}
  \caption{Computation cost (Total seconds for one time run).}
  \vspace{-6pt}
  \label{ComputationCost}
  \centering
\resizebox{0.99\columnwidth}{!}{
\begin{tabular}{c|cccc|c|c}\hline
\multirow{2}{*}{datasets} & \multicolumn{2}{c}{Emb Train} & \multirow{2}{*}{GiGaMA$_{train}$} & \multirow{2}{*}{GiGaMA$_{total}$}  & \multirow{2}{*}{GraphMAE} & \multirow{2}{*}{GCA} \\ \cline{2-3}
& PCA        & node2vec & &         &  &   \\\hline \hline
Cora    &1.22 &7.10     &8.32         &16.64        &42.38   &27.47  \\
CiteSeer   &1.12  &3.43     &6.53         &11.08         &10.33   &11.34  \\
WikiCS  &1.29      &2.78  &25.50   &29.56       &57.21      &472.08  \\   
Computers  &1.37  &14.63     & 103.31        &119.31         &62.82   &255.63  \\
Photo    &1.31  &20.45     & 121.44        &143.20         &45.60   &141.52  \\
CS  &35.58      &77.13  &247.88   &291.38        &360.59     &314.08  \\ 
Phy    &45.92 &140.23     &516.18         &702.33         &391.58   &1362.60  \\\hline            
\end{tabular}}
\vspace{-6pt}
\end{table}

\vspace{-3pt}
\subsection{Implementation Details}
All experiments are conducted on a workstation with a GPU of NVIDIA A6000. For baselines, we report the baseline model results based on their provided codes with official settings. If their settings are not available, we conduct a hyper-parameter search. The baselines are elevated under the same settings as our model on three downstream tasks. For our graph masked autoencoder, we choose GAT~\cite{velivckovic2017graph} as our encoder model and two-layer MLPs as our projector models. The GiGaMA and embedding model hyper-parameters setting is listed in Table~\ref{Student model details}.
\vspace{-3pt}
\begin{table}[h]
\renewcommand\arraystretch{1.2}
  \caption{Main hyper-parameters setting.}
  \label{Student model details}
  \centering
  \vspace{-7pt}
\resizebox{0.99\columnwidth}{!}{
\begin{tabular}{l|l|ccccccc}
\hline
&Settings &Cora  &CiteSeer &WikiCS &Computers &Photo &CS &Physics  \\ \hline
\hline
\multirow{6}{*}{GiGaMA}&Hidden dimension    &512    &256   &256      &512   &512  &512  &512 \\
&Mask\_edge  &0.4    &0.1   & 0.4     &0.4   &0.4  & 0.2 & 0.2\\
&Mask\_feature    &0.4    &0.5   & 0.4      &0.4   & 0.4 & 0.2 & 0.2\\
&$(\tilde{\lambda}_1,\tilde{\lambda}_2,\tilde{\lambda}_3)_{\mathcal{V}^E}$       &(5,2,6)    &(5,2,6)   &(5,2,6)      &(5,2,6)   &(5,2,6)  &(5,2,6)  &(5,2,6)  \\
&$(\tilde{\lambda}_1,\tilde{\lambda}_2,\tilde{\lambda}_3)_{\mathcal{V}^F}$       &(2,5,6)    &(2,5,6)   &(2,5,6)      &(2,5,6)   &(2,5,6)  &(2,5,6)  &(2,5,6)  \\
&$(\tilde{\lambda}_1,\tilde{\lambda}_2,\tilde{\lambda}_3)_{\mathcal{V}^B}$      &(1,1,3)    &(1,1,3)    &(1,1,3)       &(1,1,3)    &(1,1,3)   &(0,0,1)   &(0,0,1)   \\ \hline
PCA&Ratio    &0.5    &0.5   &0.5      &0.5   &0.5  &0.9  &0.9 \\ \hline
\multirow{4}{*}{Node2vec}&Walk length  &5    &5   &10      &10   &10  &10  &10 \\
&Context size      &5    &5   &10      &10   &10  &10  &10\\
&Walks per node     &5    &5   &10      &10   &10  &10  &10 \\
&Epoch         &20    &20   &20      &30   &100  &200  &200 \\
 \hline
\end{tabular}}
\end{table}

\vspace{-8pt}
\subsection{Theoretical Proof}
\begin{lemma} 
\textbf{Chain Rule (1).}
Given three random variables $X_1$, $X_2$, and $X_3$, we have
\begin{equation*}
    I(X_1; X_2, X_3) = I(X_1; X_3) + I(X_1 ; X_2|X_3).
\end{equation*}
\end{lemma}

\begin{proof} Using the chain rule for entropy, we first can have:
\begin{equation*}
    \begin{aligned}
    I(X_1 ; X_2 \mid X_3) & =H(X_1, X_3)+H(X_2, X_3)-H(X_1, X_2, X_3)-H(X_3) \\
    & =H(X_1 \mid X_3)+H(X_2 \mid X_3)-H(X_1, X_2 \mid X_3).
    \end{aligned}
\end{equation*}
Then the above can be re-written to
\begin{equation*}
    I(X_1 ; X_2|X_3) = I(X_1; X_2, X_3) -I(X_1; X_3).
\end{equation*}
Rearrange the above equation, we can have
\begin{equation*}
    I(X_1; X_2, X_3) = I(X_1; X_3) + I(X_1 ; X_2|X_3).
\end{equation*}
\end{proof}
\vspace{-8pt}
\begin{lemma} \textbf{Chain Rule (2).} Given three random variables $X_1$, $X_2$, and $X_3$, we have:
\begin{equation*}
I(X_1; X_2; X_3) = I(X_1; X_2) + I(X_1; X_3) - I(X_1 ; X_2, X_3).
\end{equation*}
\end{lemma}

\begin{proof} By definition, the multivariate mutual information is defined as:
\begin{equation*}
I(X_1;X_2;X_3)=I(X_1;X_2)-I(X_1;X_2 \mid X_3).
\end{equation*}
So according to Equation~(\ref{chain1}), we can get:
\begin{equation*}
    \begin{aligned}
&I(X_1; X_2 ; X_3) = I(X_1;X_2) - (I(X_1; X_2, X_3) - I(X_1; X_3))\\
& =I(X_1; X_2) + I(X_1; X_3) - I(X_1 ; X_2, X_3).
\end{aligned}
\end{equation*}

\end{proof} 

{
\bibliographystyle{ACM-Reference-Format}
\balance
\bibliography{cikm}
}

\end{document}